\documentclass[runningheads]{llncs}

\usepackage{accv}

\usepackage{accvabbrv}

\usepackage{graphicx}
\usepackage{booktabs}

\usepackage[accsupp]{axessibility}  

\usepackage{hyperref}

\usepackage{orcidlink}

\usepackage{multirow}
\usepackage{mathtools}
\usepackage{enumitem}
\usepackage{wrapfig}

\newcommand{\method}{Captioning by Discriminative Prompting}
\newcommand{\methodshort}{CDP}
\newcommand{\methodnet}{CDPNet}
\newcommand{\promptname}{discriminative prompt}

\newcommand{\g}{\textcolor[rgb]{0.0,0.5,0.0}}
\newcommand{\blue}{\textcolor[rgb]{0.0,0.3,0.8}}
\newcommand{\orange}{\textcolor[rgb]{0.8,0.3,0.0}}

\DeclareMathOperator*{\argmax}{arg\,max}

\begin{document}

\title{It's Just Another Day: Unique Video Captioning by Discriminitive Prompting} 

\titlerunning{It's Just Another Day: Unique Video Captioning.}

\author{Toby Perrett\inst{1}\orcidlink{0000-0002-1676-3729} \and
Tengda Han\inst{2}\orcidlink{0000-0002-1874-9664} \and
Dima Damen\inst{1}\orcidlink{0000-0001-8804-6238} \and
Andrew Zisserman\inst{2}\orcidlink{0000-0002-8945-8573}}

\authorrunning{Toby Perrett, Tengda Han, Dima Damen, Andrew Zisserman}

\institute{University of Bristol, UK 
\email{\{toby.perrett,dima.damen\}@bristol.ac.uk}\\ \and
University of Oxford, UK
\email{\{htd,az\}@robots.ox.ac.uk}}

\maketitle

\vspace{-15pt}
\begin{abstract}
Long videos contain many repeating actions, events and shots. These repetitions are frequently given identical captions, which makes it difficult to retrieve the exact desired clip using a text search. In this paper, we formulate the problem of unique captioning: Given multiple clips with the same caption, we generate a new caption for each clip that uniquely identifies it. We propose Captioning by Discriminative Prompting (CDP), which predicts a property that can separate identically captioned clips, and use it to generate unique captions. We introduce two benchmarks for unique captioning, based on egocentric footage and timeloop movies -- where repeating actions are common. We demonstrate that captions generated by CDP improve text-to-video R@1 by 15\% for egocentric videos and 10\% in timeloop movies.
  \begin{center}
  \url{https://tobyperrett.github.io/its-just-another-day}
  \keywords{\vspace{-10pt}Uniqueness \and Video Captioning \and Egocentric \and Movies} 
  \end{center}
\end{abstract}
\vspace{-10pt}

\section{Introduction}
\vspace{-2pt}

Life is repetitive. So videos of daily life will inevitably contain visually similar events, places, people and activities. As a consequence, captioning video clips from similar activities will often result in identical sentences. For example, in Ego4D~\cite{grauman2022ego4d} when using an off-the-shelf captioner~\cite{zhao2023learning}, 66\% of clips in each video share their caption with at least one other clip, and thus do not have a unique caption. This lack of caption uniqueness impacts text based search -- a user has to linearly scan all similar clips to find the desired clip. Can we do better?

The root of the problem is that currently clips are captioned {\em independently}~\cite{luo2020univl,seo2022end,lin2022swinbert,Yu_VideoBLIP,zhao2023learning,zhang2023video,liu2022show,lu2023set}.
Instead, if the captioner is aware of 
visually similar clips, then potentially it can discriminate one from the others in its description. 

That is the objective of this paper: to generate concise captions which discriminate between visually similar clips.
We achieve this in two ways: first, we develop a model that observes all visually similar clips, and predicts prompts that will trigger 
the captioner to generate a unique description for each. Second, if it is 
not possible to uniquely caption a clip, we increase its
temporal extent until a unique caption can be found.

In particular, we show that by taking an approach similar to the `twenty questions' game \cite{wikipedia20q},
a lightweight model can be learnt
for an already trained captioner. This allows for the direct prediction of \promptname s at inference, eliminating the need to explicitly test all possible prompts.
Our framework is agnostic to the captioner (and visual-text embedding) used.

We focus on two sources of videos with known repetitions. One is {\em egocentric footage} of
daily life, where similar clips occur naturally due to
actions and routines occurring in familiar environments. 
The other is {\em timeloop movies}, where repetition is specifically written into the plot, with the added challenge of identical or near duplicate clips.

\begin{figure}[t!]
\centering
\subfloat[A long egocentric video. The yellow/green/pink clips are captioned as ``Opens the fridge''.  \label{fig:t-egovideo}]{\includegraphics[width=0.92\textwidth]{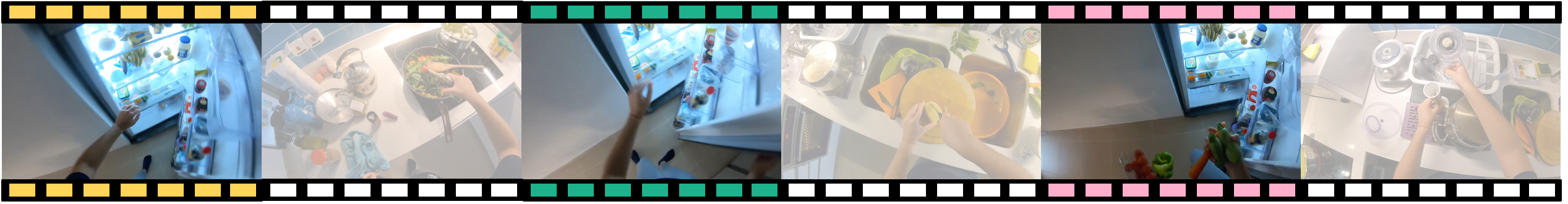}} \\ \vspace{-5pt}
\subfloat[Standard captioning can generate the same caption for multiple clips.  \label{fig:t1}]{\includegraphics[scale=0.24]{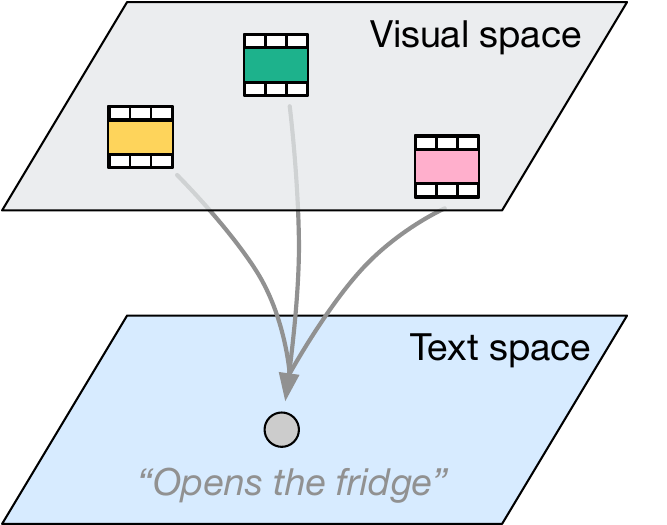}} \hspace{10pt}
\subfloat[We consider clips with the same caption, to find a property that captions them uniquely. \label{fig:t2}]{\includegraphics[scale=0.24, trim=0 0 50 0]{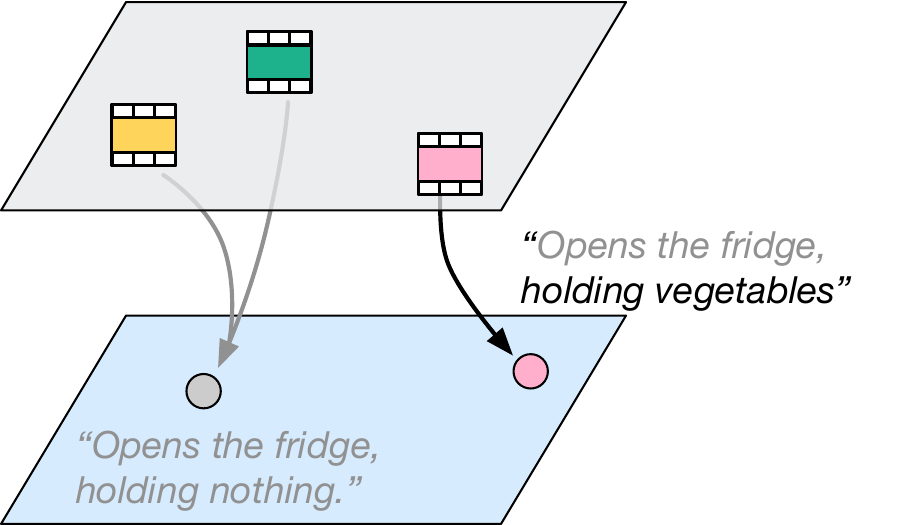}} \hspace{5pt}
\subfloat[If we cannot find a unique property, we explore the following clips for an extended unique caption. \label{fig:t3}]{\includegraphics[scale=0.24]{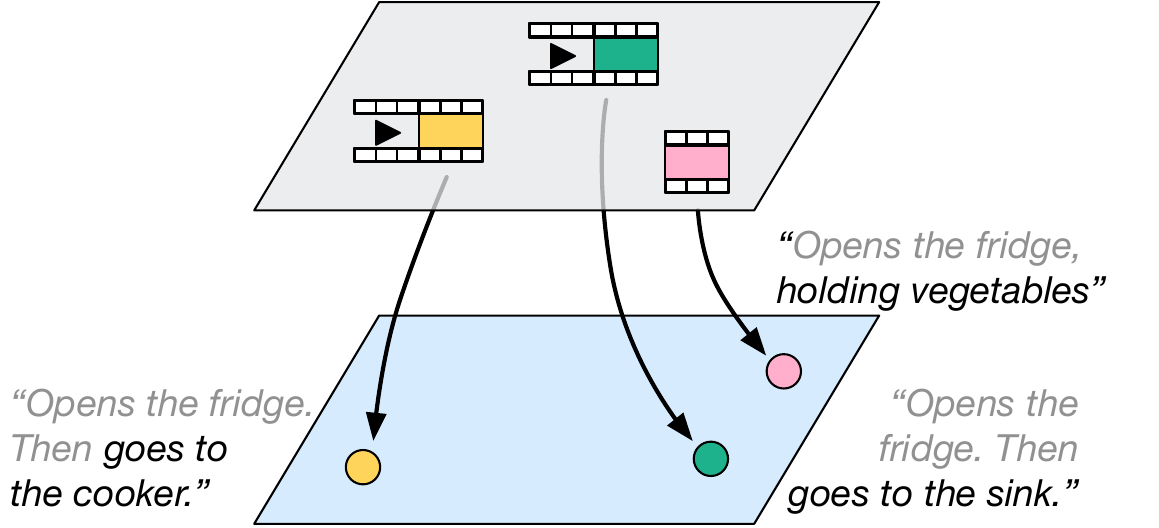}}
\vspace{-6pt}
\caption{Standard video captioning breaks the video into smaller clips and considers each clip independently. As a result, it is likely multiple clips from one video will have the same exact caption (a). We introduce \method\ (\methodshort), an approach for generating unique captions. \methodshort\ considers the set of clips with the same caption (b), and predicts a \promptname\ (\eg``holding'') that allows the clip to be captioned uniquely (c) When a unique caption cannot be found, we advance to the next clip $\blacktriangleright$ to allow unique captioning based on following actions~(d).
}
\vspace{-20pt}
\label{fig:teaser}
\end{figure}

The ability to uniquely caption video clips will enable a number of downstream tasks: 
(i) When querying for a caption in a retrieval system,
e.g.\ ``Opens the fridge'' in Fig.~\ref{fig:teaser}, the unique captions will enable an `auto-completion', appending the distinctive aspects of clips
to the text query. 
This way the user can directly select the clip they are interested in (e.g.\ ``Opens the fridge, while holding vegetables in hand'') without having to study all videos of opening the fridge.  
(ii)~Further refinement of captions on clips previously captioned identically,
without having to change the captioner.
In summary, our contributions are: 
\begin{itemize}[noitemsep,topsep=0pt,leftmargin=*]
    \item We introduce a framework for unique captioning, \method\ (\methodshort), based on predicting the dimension along which multiple clips differ. This dimension is represented by a \promptname, which conditions the captioner to generate a unique caption for each clip. 
    \item We introduce two benchmarks for unique captioning, from egocentric videos with identical narrations and from the repeating segments of timeloop movies. 
    \item We show \methodshort\ improves text-to-video and video-to-text retrieval on both benchmarks.
\end{itemize}

\vspace{-10pt}
\section{Related work}
\vspace{-5pt}
\noindent\textbf{Captioning.}
The standard practice for video captioning follows that for images \cite{mokady2021clipcap,li2022blip,li2023blip}, which is to take an auto-regressive language model, and condition it on the video \cite{luo2020univl,seo2022end,lin2022swinbert,Yu_VideoBLIP,zhao2023learning,yang2023vid2seq,zhang2023video}. Captioning models have improved with more data \cite{sharma2018conceptual,yang2023vid2seq} and better pre-trained image \cite{li2023blip} and language models \cite{zhang2023video}. Improvements have also been found by generating longer captions with more detail \cite{ding2023image}, maximising mutual information \cite{wang2020towards}, and incorporating synthetic captions \cite{betker2023improving} to challenge of the training objective. Other approaches include distilling knowledge from foundation models \cite{yang2020auto,long2023capdet,wu2023cap4video}, and alignment with human labelling \cite{sun2023aligning}. Dense \cite{yang2023vid2seq} and hierarchical \cite{islam2024video} captioning models assign multiple sequential captions to a longer clip, but do not aim for unique captions. To introduce diversity into generated captions, works have sought to produce multiple captions per clip, ensuring that the concepts contained in one caption are distinct from other captions \emph{for the same video clip} \cite{liu2022show,lu2023set}.

The problem we tackle in this paper is orthogonal to the above works. Given an already trained captioner, how can we find the differences between \emph{multiple} video clips given a captioner's existing capability? We aim to generate concise, unique captions for all clips in a video, or a gallery of videos. 

\vspace*{6pt}

\noindent\textbf{Model ``blind spots''.}
A related line of work is determining aspects of an image or video that models are blind to, and thus unable to tell apart, typically evaluated on classification and VQA tasks. Examples include compositionality \cite{thrush2022winoground} in images, temporal ordering in video \cite{hendricks2018localizing}, and the limitations of image/language pre-trained models compared to self-supervised image representations \cite{tong2024eyes}.
These works have aimed at evaluating and providing benchmarks, but some also attempt to fix these shortcomings, for example by collecting data focusing on these blind spots \cite{bagad2023test,ventura23covr}, or training captioners on LLM processed auto-labels to find differences between pairs of samples \cite{nagarajan2024step,lin2024comparison}. 
In this work, instead of trying to improve the base ability of a captioning model, we acknowledge that limitations will likely always exist. We find differences which are discernible by the chosen existing captioning model. 
This approach will continue to be relevant and useful as models' capabilities continue to improve.

\vspace*{6pt}

\noindent\textbf{Long video.}
Early computer vision studies of long video focused on footage from surveillance \cite{hampapur2005smart}, TV \cite{duan2006segmentation,xu2003video} and movies \cite{sivic2003video}. Due to computational budgets, the community shifted to understanding short clips \cite{carreira2017quo,goyal2017something}. Long-video is again being studied, due to the rise of large-scale instructional \cite{zhou2018towards,miech2019howto100m} and egocentric datasets~\cite{damen2022rescaling,grauman2022ego4d}, and the ability of models to operate on larger temporal windows~\cite{wu2022memvit,wang2022language,liu2023ring}.
Several studied tasks in long video would benefit from the ability to uniquely caption clips, such as summarisation \cite{lee2012discovering}, audio description \cite{han2023autoad}, VQA~\cite{tapaswi2016movieqa} and retrieval \cite{ramakrishnan2023spotem}.
While visual-language retrieval benchmarks~\cite{rohrbach2015lsmdc,xu2016msrvtt} report Recall@K accuracy as a common practice to allow retrieving similar instances in the corpus, they do not explicitly enforce uniqueness.
In this paper, we build our egocentric unique captioning benchmark using footage from the massive-scale Ego4D dataset~\cite{grauman2022ego4d}. 

\noindent\textbf{Timeloop movies.}
Despite the challenges they pose to plot understanding, timeloop movies rarely feature in the computer vision literature, where the emphasis for movies/TV is on scale~\cite{lei2018tvqa,bain2020condensed,huang2020movienet,yue2023movie101}. Whilst they are scarce compared to standard movies (with only 71 dating back to 1947 listed on Wikipedia \cite{wikipediatimeloops}) and thus not suitable for large-scale training, timeloop movies can provide insightful diagnostic and qualitative results. For example, the movies
\emph{Groundhog Day} and \emph{Run Lola Run} were used for location retrieval assessment in~\cite{sivic2003video}, as tests of identical and near-duplicate shot retrieval in~\cite{chum2007scalable}, and the movie's repeating structure
was determined in~\cite{Schaffalitzky03} by matching shots of the same location. 
In this work, we revisit timeloop movies as an evaluation-only benchmark for unique video captioning, where understanding the temporal context of repetitive clips is essential.

\vspace{-10pt}
 \section{Method: \method}
\vspace{-5pt}

\method\ (\methodshort), generates {\em unique captions}
for a set of visually similar clips, so they can be discriminated and teased apart in the visual-text embedding space.
\methodshort\ is built around three key ideas:
\begin{enumerate}[noitemsep,topsep=0pt,leftmargin=*]
    \item  A set of \emph{discriminative prompts}, in order to direct a captioner to focus on properties of one clip which distinguish it from others. These properties are chosen by contrasting all similar clips, and thus provide our mechanism for conditioning a single-clip captioner on multiple clips.
    \item A \emph{combinatorial search} over all prompts and clips, to find the exact combination of prompts that will generate the most unique set of captions.
    \item A network, \emph{\methodnet}, which approximates the most computationally expensive part of the search - auto-regressively captioning each clip with all prompts and then computing video/text embedding similarities.
\end{enumerate}
Note that we can, with significant computational cost, generate unique captions by performing the combinatorial search for \promptname s without training an additional network. This would be possible but inefficient. We introduce this process first in Section~\ref{sec:combinatorial} as our method builds on the search, and it provides a good insight into the constraints and comparisons necessary for caption uniqueness. \methodnet\ is a required approximation to make our proposed approach feasible for inference, described in Section~\ref{sec:predprompts}.

\vspace{-12pt}
\subsection{Problem statement and uniqueness definition}\label{sec:probdef}
\vspace{-5pt}

Given a set of $N$ video clips $\mathcal{V} = \{v_1, v_2,...,v_N\}$, we aim to output a corresponding set of unique captions $\mathcal{C} = \{c_1,c_2,...,c_N\}$.
We assume access to two trained foundation models:
\begin{itemize}[noitemsep,topsep=0pt,leftmargin=*]
    \item A video captioner $\Theta(v, p)$, which takes in a video clip $v$ and optional prompt $p$, and produces the caption text $c$.
    \item A dual-encoder video/text model,  with encoders $f(v)$ and $g(c)$ for projecting video clips and caption text into 
a joint embedding space. The video-text similarity is measured by cosine similarity between their embeddings.
\end{itemize}
These trained models are general and do not need to have been trained jointly. They remain frozen throughout.

A correct output of our method is one where captions are distinct, i.e. ${c_i \ne c_j; \forall i \ne j}$, but also can be correctly matched to the clip it was generated from. 
Formally, $v_i$ is captioned uniquely by $c_i$ with respect to all other $v_j \in \mathcal{V}$ and $c_j \in \mathcal{C}$ if the following condition is satisfied, where $\langle\cdot, \cdot\rangle$ denotes cosine similarity:
\vspace{-6pt}
\begin{equation} \label{eq:conditions}
\langle f(v_i), g(c_i)\rangle > \max \Bigl(\max_{j \neq i} \langle f(v_i), g(c_j)\rangle, \max_{j \neq i} \langle f(v_j), g(c_i)\rangle\Bigr).
\vspace{-6pt}
\end{equation}

\subsection{Combinatorial search for unique captions}\label{sec:combinatorial}
\vspace{-5pt}
\noindent \textbf{Discriminative prompts. }
For each clip $v_i$, our
goal is to select one or more \promptname (s)  that will induce a
unique caption with respect to all other clips in~$\mathcal{V}$. 
While there are many ways in which clips can differ, we propose to use a given set of general prompts.
We define a bank of $P$ prompts ${\mathcal{B}=\{p_1,...,p_P\}}$.  Fixed prompts are more suited to this task than learned, as they are interpretable, and can be designed to increase diversity and reflect known model capabilities. We typically select the most frequent N-grams from the training set.

\noindent\textbf{Selecting a single \promptname.}
We define the similarity function, $s$, between a video $v_i$ and a caption generated from another video $v_j$ using the prompt $p_k \in \mathcal{B}$ as \vspace{-10pt}
\begin{equation} 
    s (v_i, v_j, p_k) = \langle f(v_i), g(\Theta(v_j, p_k)) \rangle,
\label{eq:similarity1}
\end{equation}
which measures clip/caption similarity in the shared embedding space.
We next define the uniqueness margin, $\mathcal{M}$, for clip $v_i$, with respect to the other clips in $\mathcal{V}$, using prompt $p_k$, as \vspace{-5pt}
\begin{equation}
         \mathcal{M}(v_i, p_k) = s(v_i, v_i, p_k) - \max \Bigl(\max_{j \neq i} \bigl(s(v_j, v_i, p_k)\bigr), \max_{j \neq i} \bigl(s (v_i, v_j, p_k)\bigr) \Bigr).
         \label{eq:margin}
\end{equation}
For the clip $v_i$, we denote the chosen \promptname\ as~ $p_{\hat{k}}$, which is the prompt with index $\hat k$ that maximises $\mathcal{M}$, \ie ${\hat{k}= \argmax_k \mathcal{M}(v_i, p_k)}$. 
If $\mathcal{M}(v_i, p_{\hat{k}}) > \lambda$, where $\lambda$ is the margin of confidence, then the caption  $c_i = \Theta(v_i, p_{\hat{k}})$ uniquely describes $v_i$, as defined by Eq. \ref{eq:conditions}. That is,  $v_i$ is closer to its caption $c_i$ than to any other caption, and $c_i$ is closer to $v_i$ than to any other video.
If $\mathcal{M}(v_i, p_{\hat{k}})\leq \lambda$, then we have determined that it is not possible to identify $v_i$ uniquely, given the other video clips in the set and the capabilities of the captioner and embedding space. 
In such cases, multiple \promptname s are required to find unique captions, which we describe next.

\noindent\textbf{Selecting multiple \promptname s.}
Given $P$ prompts in our bank and $N$ clips in $\mathcal{V}$, the exact search for the best combination of \textit{multiple} prompts would have complexity $\mathcal{O}(NP^P)$ if every prompt could be chosen for every clip. Instead, we constrain the maximum number of prompts to be selected for each clip as $\alpha << P$, which reduces the complexity to $\mathcal{O}(NP^\alpha)$.
This not only constrains the search space, but also keeps our captions concise and focused on the most discriminative aspects of each clip.

Our problem is now to define the margin for combinations of prompts, and to find the combination which obtains the maximum margin.
The margin for a single prompt reflects that a clip is distinct from all others with respect to that single prompt.
We extend to multiple prompts by adjusting the similarity measure (Eq.~\ref{eq:similarity1}) used to compute the margin. Instead, we take the mean of the similarities of each individual prompt in the combination.

We first define all combinations of prompts in $\mathcal{B}$ up to order $\alpha$ as $\mathcal{B}^\alpha$. For example, $\mathcal{B}^2$ contains all the combinations of $\{p_i\}$ and $\{p_i,p_j\}$. We denote the k-th combination in $\mathcal{B}^{\alpha}$ as $ {B}^{\alpha}_k$. We then adjust the similarity to:
\vspace{-6pt}
\begin{equation}
    s^\alpha (v_i, v_j, B_k^{\alpha}) = \frac{1}{|B_k^\alpha|} \sum_{p \in B_k^\alpha} \langle f(v_i), g(\Theta(v_j, p))\rangle
    \label{eq:similarity2}
\vspace{-6pt}
\end{equation}
We use $s^\alpha$ instead of $s$ in calculating the uniqueness margin (in Eq. \ref{eq:margin}) to choose the best prompt combination $B_{\hat{k}}^{\alpha}$.

\begin{figure}[t]
\centering
\subfloat[Each clip is captioned using every prompt. \label{fig:comb_pipeline_caption}]{\includegraphics[scale=0.24,trim = 0 0 0 5]{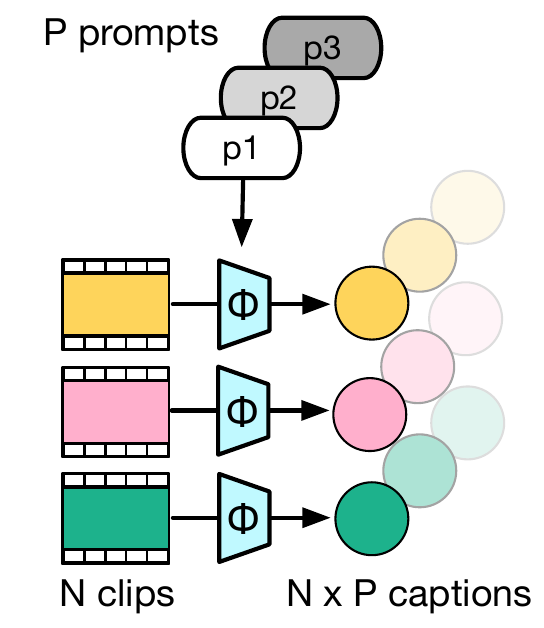}} \hspace{10pt}
\subfloat[Clips and captions are projected by $f$ and $g$ into a shared space, where \blue{similarities}  (Eq. \ref{eq:similarity2}) are computed.  \label{fig:comb_pipeline_similarities}]{\includegraphics[scale=0.24,trim = 0 0 0 5]{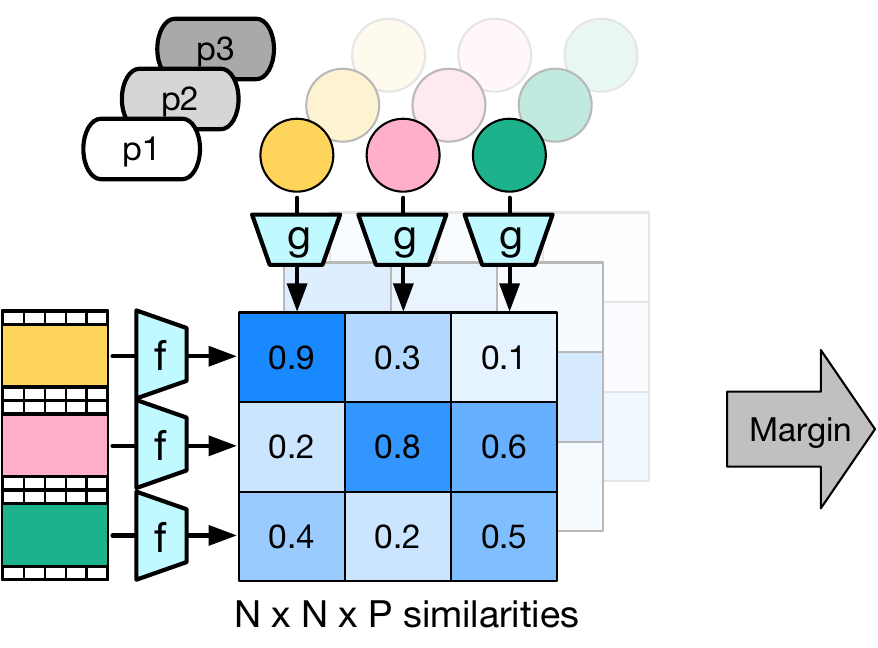}} \hspace{10pt}
\subfloat[The \orange{margins} (Eq. \ref{eq:margin}) are computed using the \blue{similarities} for every clip/prompt combination. We search for the max margin per clip, which generates a unique caption if $> \lambda$.  \label{fig:comb_pipeline_margin}]{\includegraphics[scale=0.24, trim=0 -10 0 5]{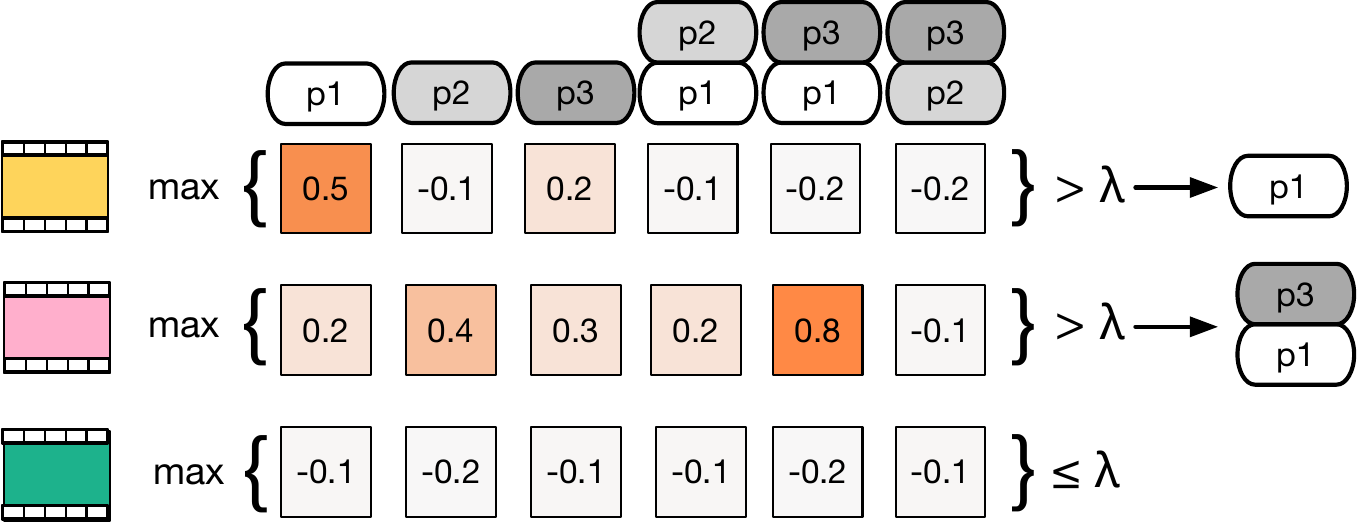}} 
\vspace{-15pt}
\caption{Pipeline for computing the margin at a single timestep for three clips. This example uses $\alpha=2$, so margins are computed for all single prompts and pairs of prompts. (a) and (b) are replaced by a learned network in Sec. \ref{sec:predprompts}.}
\vspace{-15pt}
\label{fig:comb_pipeline}
\end{figure}

We now have a complete pipeline for computing margins, and thus creating unique captions, for a set of clips with no learning necessary, which is illustrated in Fig. \ref{fig:comb_pipeline}. In \ref{fig:comb_pipeline_caption}, captions are extracted with every prompt for every clip using the frozen single-clip captioner. In \ref{fig:comb_pipeline_similarities}, clips and captions are projected into the embedding space by $f$ and $g$, where their similarities are computed. In \ref{fig:comb_pipeline_margin}, for each video, margins are computed from the similarities for every prompt combination (here, $\alpha=2$, so all individual and pairs of prompts are tried). If the maximum margin for a clip is $> \lambda$, \ie $\mathcal{M}(v_i, B_{\hat{k}}^{\alpha}) > \lambda$, then the prompt combination with the maximum margin generates a unique caption for that clip. 
However, it is possible that no combination of prompts is able to uniquely identify a clip, i.e. $\mathcal{M}(v_i, B_{\hat{k}}^{\alpha}) \leq \lambda$. In such cases, we allow the temporal duration of video clips to be extended in time, which we describe next.

\noindent\textbf{Temporal extension}
As clips are taken from a longer video, we explore advancing to a subsequent clip so as to distinguish identical clips.
As a result of the expansion, we can caption the two clips into ``X then Y'' vs ``X then Z'', where Y and Z are distinct follow-up events in the longer video.
We denote a video $v_i$ advanced by time $t$ as $v_i(\blacktriangleright t)$. We define the set of all (prompt, time) combinations up to time $\tau$ as $\mathcal{B}^\tau$ and define the $k$-th element as $B^{\tau}_{k}$. Our problem is now to adjust the similarity to account for prompt/time combinations. We can again achieve this by adjusting the similarity for clips up to time $\tau$ as follows:
\vspace{-6pt}
\begin{equation}
    s^\tau (v_i, v_j, B_k^\tau) = \frac{1}{|B_k^\tau|} \sum_{p, t \in B_k^\tau}  \langle f(v_i(\blacktriangleright t)), g(\Theta(v_j(\blacktriangleright t), p))\rangle
    \label{eq:similarity3}
\vspace{-6pt}
\end{equation}
Similar to the case of multiple prompts at a single timestep above, we replace $s$ in Eq. \ref{eq:margin} with $s^\tau$. The process for generating unique captions is then the same as Fig.~\ref{fig:comb_pipeline}. The only differences would be the tensor in \ref{fig:comb_pipeline_similarities} having an additional dimension for the number of timestep advances (\ie $N \times N \times P \times \tau$), and additional margin computations for prompt combinations at different timesteps.

\vspace{-12pt}
\subsection{Predicting \promptname s with \methodnet}\label{sec:predprompts}
\vspace{-5pt}

In Sec.~\ref{sec:probdef} and \ref{sec:combinatorial}, we predict the best prompts exhaustively and exactly. That is, we have an exact process for generating unique captions in a given video/text space, with no training or fine-tuning. It uses exact embeddings of captioner outputs, computes margins, and searches over the set of prompt combinations to find the optimal combination. However, this process is not scalable. We recognise that the main bottleneck is attempting to generate a caption for every one of the $N \times P \times \tau$ video/prompt/time combinations.

We approximate this by training a Captioning by Discriminative Prompting Network, \methodnet, denoted $\Psi$. 
It takes in two clips $v_i$ and $v_j$ and a prompt $p_k$. It predicts the video-text similarity between the visual embedding of $v_i$, and the text embedding of $v_j$ when captioned using prompt $p_k$. We denote this predicted similarity as:
\begin{equation} \label{eq:Psi}
  \hat{s}_{ijk} = \Psi(v_i, v_j, p_k).
\end{equation}
Importantly, this allows us to replace Fig. \ref{fig:comb_pipeline_caption} and \ref{fig:comb_pipeline_similarities} with the direct prediction of similarities from video clips only, \emph{without having to calculate a forward pass through the captioner or the embedding networks}.

We train $\Psi$ by minimising the MSE between its output $\hat{s}_{ijk}$, and the computed cosine similarity of the embeddings $s(v_i, v_j, b_k)$ from Eq. \ref{eq:similarity1}. To allow scalability, CDPNet $\Psi$ only operates on one prompt, and the search for combinations of prompts is handled by the averaging of similarities in the combinatorial search~(Eq. \ref{eq:similarity3}).

\begin{figure}[t]
\centering
\subfloat[A frozen captioner is run on a video input with a \promptname\  from the bank.\label{fig:m1}]{\includegraphics[scale=0.26,trim= 0 0 0 5]{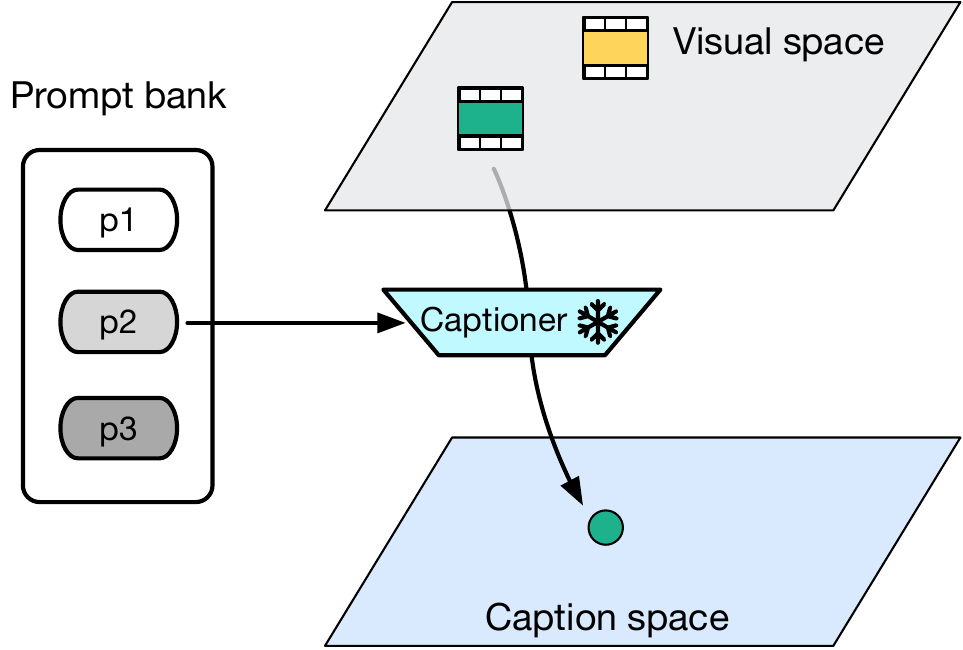}} \hspace{10pt}
\subfloat[$f$ and $g$ project clips and captions, where their similarities are computed. \label{fig:m2}]{\includegraphics[scale=0.26,trim= -20 0 -20 5]{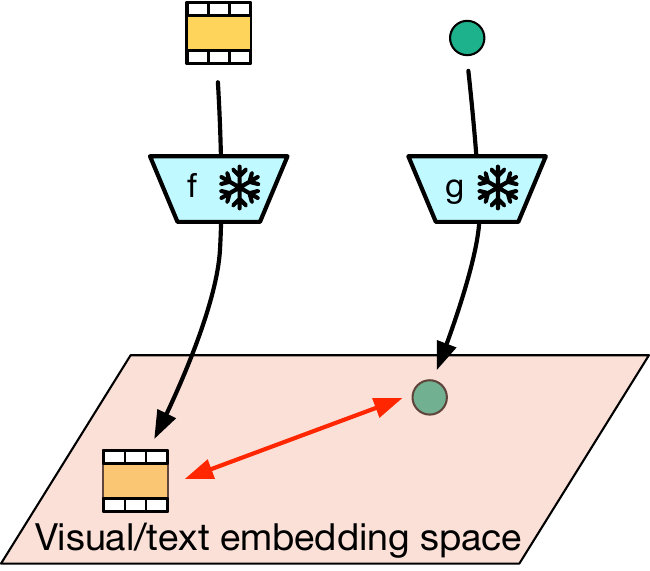}} \hspace{10pt}
\subfloat[CDPNet estimates embedding similarities.\label{fig:m3}]{\includegraphics[scale=0.26,trim= 0 0 0 5]{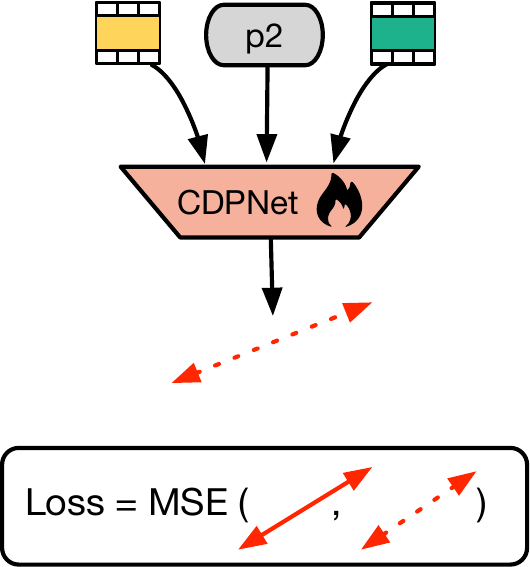}}
\vspace{-8pt}
\caption{Training \methodnet, which aims to predict the similarity between a clip (yellow) and the caption from another clip (green), when conditioning the captioner with a prompt. (a) and (b) show how to compute the similarity between the clip and caption in a shared embedding space, which is used as the training signal. (c) shows \methodnet\ predicting the similarity only using the video clips and prompt in one forward pass.}
\vspace{-12pt}
\label{fig:method}
\end{figure}

Fig. \ref{fig:method} shows the training process for \methodnet: \ref{fig:m1} shows a green clip, and its generated caption (green circle), when captioned using a prompt from the bank; \ref{fig:m2} shows the yellow clip and green caption being embedded by $f$ and $g$ in the shared visual/text space, where their similarity is computed; and \ref{fig:m3} shows how this similarity is used as the training signal for $\Psi$, which takes just the clips and prompt (\ie no caption).

We implement $\Psi$ as a transformer encoder, and its prompt bank as a collection of learnable tokens, one per prompt. $\Psi$ takes as input representations for the two clips as well as the prompt token selected from the bank. We use learned positional encodings to indicate the different inputs. We apply a linear layer to the output of the prompt token to regress the similarity $\hat{s}$.

In summary, \methodnet\ operates as follows. It is initialised with a bank of prompts. For a given set of clips, \methodnet\ directly predicts the visual-text similarity between each clip and every other clip with respect to every prompt in the bank, using visual input only (\ie no captioning). The combinatorial search is run over \methodnet\ similarity predictions to find the combination of (max $\alpha$) prompts for each clip which produces the maximum margin. If this margin is $> \lambda$, the clip and prompt combination are passed to the captioner to expecting unique captions. If the margin is $\le \lambda$, the clip is advanced ($\blacktriangleright$) and the process is repeated until a unique prompt combination is found.

\vspace{-10pt}
\section{Unique captioning benchmarks} \label{sec:benchmark}

\begin{figure}[t]
    \centering
    \includegraphics[width=0.98\textwidth]{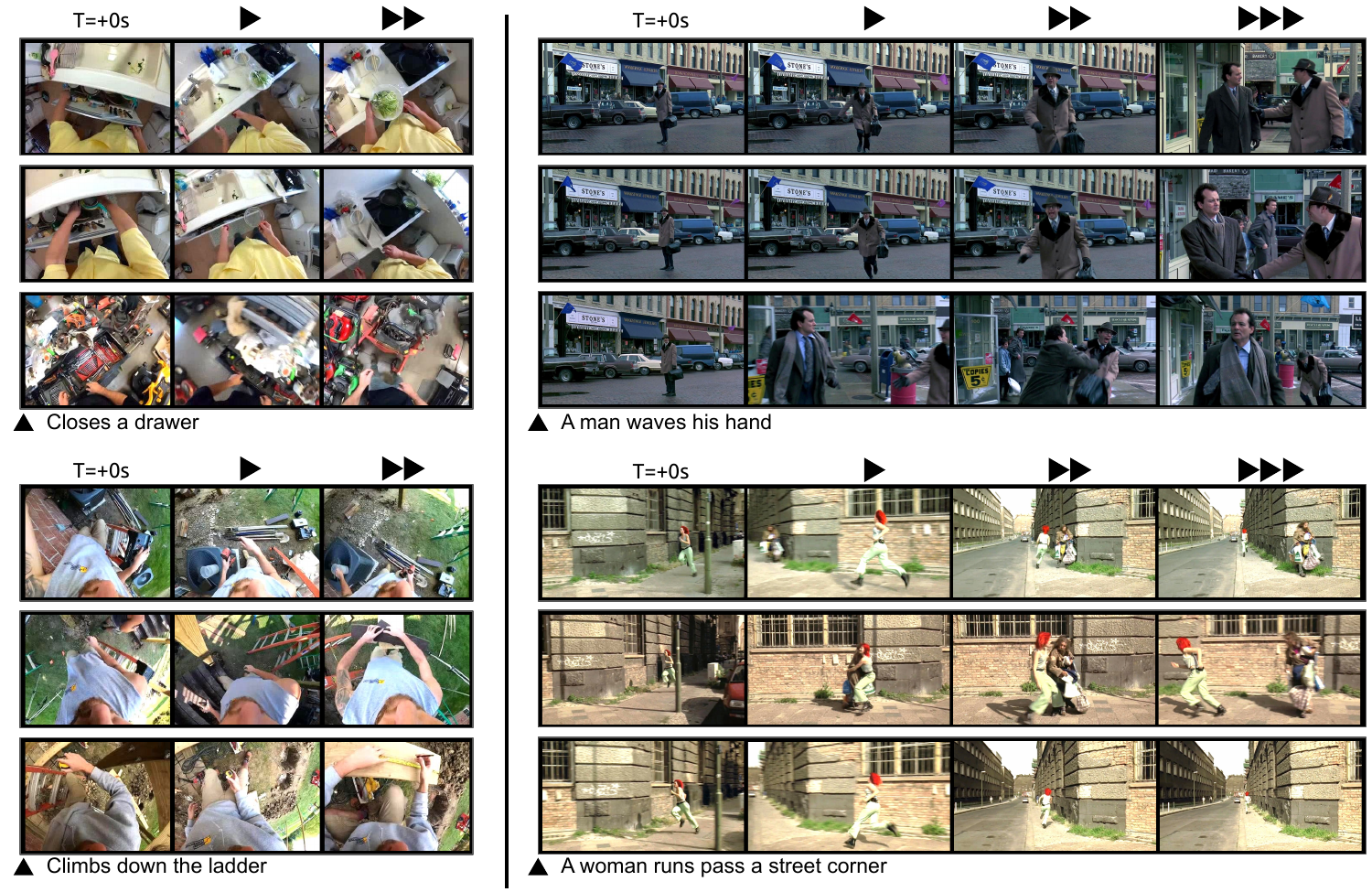}
    \vspace*{-10pt}
    \caption{Examples of the Unique Captioning Benchmarks, from Egocentric videos (left) and timeloop movies (right). We show 3 sequences from each set of clips -- i.e.\ video clips with the same caption at T=+0. Subsequent clips are indicated by $\blacktriangleright$. We note the common caption in each case. }
    \vspace*{-15pt}
    \label{fig:benchmark}
\end{figure}
\vspace{-10pt}

The ability to uniquely caption video clips can be assessed by checking if there is a one-to-one correspondence between the set of video clips and the predicted set of captions. We frame this as a retrieval problem and use standard retrieval metrics accordingly. 

\noindent\textbf{Metrics.}
Given a set of video clips $\mathcal{V}$ and generated captions $\mathcal{C}$, we measure how well each caption $c_i$ can retrieve the video $v_i$ and vice versa. We use both text-to-video and video-to-text recall @1, @2 and @3. Text-to-video R@K indicates that $v_i$ is one of the K-closest videos to $c_i$ in the shared embedding space, and vice versa. We present Avg R@1 as a standard combined metric. However, Avg R@1 does not check for one-to-one assignment. Accordingly, we introduce the more stringent Cycle@1 combined metric. This checks that video clip $v_i$ retrieves the corresponding $c_i$ \emph{and} $c_i$ then retrieves $v_i$. Formally, this means $v_i$ and $c_i$ must satisfy the uniqueness condition in Eq. \ref{eq:conditions}.

\noindent\textbf{Egocentric benchmark.}
Videos of daily living are an ideal test of unique captioning, as we perform the same activities and routines over both short and long timescales. The egocentric viewpoint naturally captures these repetitions as we move around, so we use videos and narrations from Ego4D to construct this benchmark. For training, 
We sample 30K ground truth narrations from the Natural Language Query (NLQ) training split that are repeated 10 or more times, i.e.\ 10 or more clips share the same narrations.
For each, we randomly sample 10 clips from the same or a different video, producing a benchmark of 300K clips.
Our balanced sampling, per narration, helps alleviate the long-tail property where some narrations such as ``look around'' are common. 

For evaluation, we select 300 sets of 10 clips from the NLQ validation set, following the same process. Sets remain fixed for all evaluations.
To select prompts for the bank $\mathcal{B}$, we take N-grams from narrations on the training set. We sort by N-gram frequency, selecting the 10 largest, whilst manually removing semantically similar prompts to ensure diversity. Prompts are ablated in Sec. \ref{sec:ablation_bank}. 

\noindent\textbf{Timeloop movies benchmark.}
A timeloop movie is one where the plot centres around the characters reliving the same sequence of events over and over. For example, in the 1998 movie `Run Lola Run', the character repeatedly starts running from the same premise in her repeated attempts to save her boyfriend. At each repetition, a difference occurs, which causes the story to diverge. This makes these timeloop movies an ideal test for unique captioning.

For the test set, we start with the Wikipedia timeloop list \cite{wikipediatimeloops},
and manually annotate the timestamps of these repeating moments.  We only select sets when there are at least 3 repeating moments per movie, and where these moments are near-identical, making this a very challenging evaluation.
For example, in the movie \emph{Groundhog Day}, the main character is clearly filmed waking up 9 times, from which the story diverges. 
These ``waking-up'' clips form a set of 9 clips.
This process gives a total of 63 clips across 10 timeloop movies, ranging from 1993 to 2021. Sets range in size from 3 to 10.

Due to the scarcity of timeloop movies, they can only be used for a test set. For training , we use non-timeloop data as a proxy for timeloop clips. We curate our training set from the Condensed Movie Dataset~\cite{bain2020condensed}, which consists of sequences of 2-minute movie truncations, from which we \emph{exclude timeloop movies}. We construct sets of 10 video clips with CLIP visual similarity > 0.92 (chosen to match the egocentric training size), giving 30K sets of 10 clips. 
Examples from both benchmarks are shown in Fig. \ref{fig:benchmark}.

Our benchmark and training clips are available from the project's website.

\vspace{-10pt}
\section{Experiments}
\vspace{-6pt}
\subsection{Baselines}
\vspace{-6pt}

Our proposed method, \method\ (\methodshort), is applicable to any captioning model and does not fine-tune the captioner. 
Here, we focus our experiments on the SOTA baseline model for each benchmark, with other models in the supplementary \cite{Yu_VideoBLIP,yu2023efficient,zhang2023video,zhao2023learning}.
For egocentric, we use the Visually-Conditioned-Language-Model (VCLM) from LaViLa~\cite{zhao2023learning}, which is specifically trained to caption egocentric clips. For the timeloop movie benchmark, we use Video-LLaMA \cite{zhang2023video}, which has recently been successfully used for describing movies~\cite{song2023moviechat,han2024autoad}.
We use publicly available captioner checkpoints. 
Combined with our \methodshort, we demonstrate improved performance in every case.

\methodshort\ advances to the next clip in the continuous video when a unique caption cannot be predicted. For direct comparison,  we evaluate all models on the same number of clips. 
$T = +0s$ indicates the models can only see one clip, which is 5s for egocentric and 2s for timeloop following model defaults. For the egocentric benchmark, $T = +5s$ indicates the methods are allowed access to the next $5s$ clip, $T = +10$ indicates access to the next two clips and so on. The timeloop movie benchmark, the equivalents are $T= +2s$ and $T = +4s$.

\vspace{-10pt}
\subsection{Implementation details.}
\vspace{-5pt}

We implement \methodnet\ ($\Psi$ in Eq. \ref{eq:Psi}) as a transformer encoder, with 2 layers, 4 heads and feedforward dimension of 1024. It is trained for 25 epochs using Adam with lr 0.0001, decaying by a factor of 10 at epochs 15 and 20, and batch size of 64. 
Default hyperparameter values are $\alpha=3$ (Eq. \ref{eq:similarity2}) and $\lambda=0.1$ (Sec. \ref{sec:predprompts}).

For the egocentric benchmark, we use EgoVLP 336px \cite{qinghong2022egocentric} as the evaluation network. For the captioner, we take the LaViLa VCLM-HR \cite{zhao2023learning}, which is a GPT-2XL with trained cross attention layers and bridge to the visual features, and the Timesformer-L/Distilbert-Base 256d visual/text embedding network for training. Video clips are 5s long, from which 4 frames are uniformly sampled (the default). At maximum temporal extension, the inclusion of \methodnet\ and the search only increases captioning time from 4.5 to 5.8s, compared to 300s when we do not use \methodnet\ searching exhaustively.

For timeloop movies, we use InternVideo \cite{wang2023internvid} as the video/text evaluation space with 8$\times$224px frames projected to a 768d feature. We use Video-LLaMA~\cite{zhang2023video} as the captioner, operating on 8$\times$224px uniformly sampled frames, and EVA-CLIP~\cite{sun2023eva} video/text features as the embedding space during training.

\vspace{-10pt}
\subsection{Results}
\vspace{-5pt}

\begin{table}[t]
\centering
\resizebox{0.85\linewidth}{!}{
\begin{tabular}{@{}lclcccccc|ll@{}}
\toprule
 &  &  & \multicolumn{3}{c}{\textbf{Text$\rightarrow$Video}} & \multicolumn{3}{c}{\textbf{Video$\rightarrow$Text}} &  &  \\ 
\textbf{T} & \textbf{\#Clips} & \textbf{Method} & \textbf{R@1} & \textbf{R@2} & \textbf{R@3} & \textbf{R@1} & \multicolumn{1}{c}{\textbf{R@2}} & \textbf{R@3} & \textbf{Avg R@1} & \textbf{Cycle@1} \\ \midrule 
 &  & {\color[HTML]{999999} Chance} & {\color[HTML]{999999} 10} & {\color[HTML]{999999} 20} & {\color[HTML]{999999} 30} & {\color[HTML]{999999} 10} & {\color[HTML]{999999} 20} & {\color[HTML]{999999} 30} & {\color[HTML]{999999} 10} & {\color[HTML]{999999} 1.0} \\ \midrule
\multirow{2}{*}{+0s} & \multirow{2}{*}{1} & LaViLa VCLM & 40 & 58 & 70 & 33 & 49 & 61 & 37 & 22.0 \\ \vspace{1pt}
 &  & LaViLa VCLM + \methodshort & \textbf{55} & \textbf{71} & \textbf{80} & \textbf{34} & \textbf{50} & \textbf{61} & \textbf{45 }\g{  +8} & \textbf{26.0 } \g{  +4.0} \\ \midrule
\multirow{2}{*}{+5s} & \multirow{2}{*}{2} & LaViLa VCLM & 42 & 61 & 72 & 34 & 51 & 62 & 38 & 23.0 \\ \vspace{1pt}
 &  & LaViLa VCLM + \methodshort & \textbf{69} & \textbf{81} & \textbf{88} & \textbf{44} & \textbf{60} & \textbf{71} & \textbf{57 }\g{  +19} & \textbf{38.6 }\g{  +15.6} \\ \midrule
\multirow{2}{*}{+10s} & \multirow{2}{*}{3} & LaViLa VCLM & 45 & 63 & 74 & 36 & 52 & 64 & 41 & 25.3 \\ \vspace{1pt}
 &  & LaViLa VCLM + \methodshort & \textbf{77} & \textbf{87} & \textbf{92} & \textbf{53} & \textbf{68} & \textbf{77} & \textbf{65 }\g{  +24} & \textbf{47.1 }\g{  +21.8} \\ \midrule
\multirow{2}{*}{+30s} & \multirow{2}{*}{7} & LaViLa VCLM & 47 & 66 & 76 & 38 & 55 & 67 & 43 & 27.2 \\ \vspace{0pt}
 &  & LaViLa VCLM + \methodshort & \textbf{86} & \textbf{92} & \textbf{95} & \textbf{66} & \textbf{80} & \textbf{85} & \textbf{76 }\g{  +33} & \textbf{62.3 }\g{  +35.1} \\ \bottomrule
\end{tabular}
}
\caption{Egocentric benchmark using the LaViLa VCLM as the base captioner, and combined with  \method\ (\methodshort). At every T, \methodshort\ improves by a significant margin on every metric. Improvements are shown in \g{green} for the combined metrics Avg R@1 and Cycle@1. 
}
\vspace*{-15pt}
\label{tab:resultsego}
\end{table}

\begin{table}[t]
\centering
\resizebox{0.85\linewidth}{!}{
\begin{tabular}{@{}lclcccccc|ll@{}}
\toprule
 &  &  & \multicolumn{3}{c}{\textbf{Text$\rightarrow$Video}} & \multicolumn{3}{c}{\textbf{Video$\rightarrow$Text}} &  &  \\ 
\textbf{T} & \textbf{\#Clips} & \textbf{Method} & \textbf{R@1} & \textbf{R@2} & \textbf{R@3} & \textbf{R@1} & \textbf{R@2} & \textbf{R@3} & \textbf{Avg R@1} & \textbf{Cycle@1} \\ \midrule
 &  & {\color[HTML]{999999} Chance} & {\color[HTML]{999999} 16} & {\color[HTML]{999999} 32} & {\color[HTML]{999999} 48} & {\color[HTML]{999999} 16} & {\color[HTML]{999999} 32} & {\color[HTML]{999999} 48} & {\color[HTML]{999999} 16} & {\color[HTML]{999999} 3.0} \\  \midrule
 &  & Video-LLaMA & 37 & 65 & 84 & 34 & 54 & 59 & 35 & 18.3 \\ \vspace{1pt}
\multirow{-2}{*}{0s} & \multirow{-2}{*}{1} & Video-LLaMA + \methodshort & \textbf{47} & \textbf{71} & \textbf{86} & \textbf{36} & \textbf{63} & \textbf{78} & \textbf{42} \g{+7} & \textbf{25.0} \g{+6.7} \\ \midrule
 &  & Video-LLaMA & 55 & 70 & 82 & 32 & 64 & 75 & 43 & 25.4 \\ \vspace{1pt}
\multirow{-2}{*}{2s} & \multirow{-2}{*}{2} & Video-LLaMA + \methodshort & \textbf{51} & \textbf{70} & \textbf{81} & \textbf{45} & \textbf{64} & \textbf{75} & \textbf{48} \g{+5} & \textbf{32.0} \g{+6.6} \\ \midrule
 &  & Video-LLaMA & 53 & 70 & 83 & 33 & 56 & 75 & 43 & 18.4 \\ \vspace{1pt}
\multirow{-2}{*}{4s} & \multirow{-2}{*}{3} & Video-LLaMA + \methodshort & \textbf{62} & \textbf{77} & \textbf{85} & \textbf{44} & \textbf{68} & \textbf{84} & \textbf{53} \g{+10} & \textbf{37.4} \g{+19.0} \\ \midrule
 &  & Video-LLaMA & 44 & 70 & 83 & 32 & 56 & 74 & 38 & 18.2 \\ \vspace{1pt}
\multirow{-2}{*}{10s} & \multirow{-2}{*}{5} & Video-LLaMA + \methodshort & \textbf{73} & \textbf{87} & \textbf{95} & \textbf{53} & \textbf{75} & \textbf{84} & \textbf{63} \g{+25} & \textbf{44.5} \g{+26.3} \\ \midrule
\end{tabular}
}
\caption{Timeloop movie benchmark using Video-LLaMA as the base captioner. 
\methodshort\ is able to pick out more differences as the storylines diverge.}
\vspace{-24pt}
\label{tab:resultstimeloop}
\end{table}

Table \ref{tab:resultsego} shows results on the egocentric benchmark. At every timestep in every metric, \methodshort\ outperforms the captioner, LaViLa VCLM. 
The improvement is 8\% Avg R@1 and 4\% Cycle@1 acting on just the 5s clips without any advancement (\ie $T = +0$). 
Whilst the LaViLa VCLM obtains small improvements with access to subsequent clips (\ie as T increases), as expected, it is not able to pick out the specific aspects which make each clip unique. In contrast, the mechanism in \methodshort\ to find uniqueness provides greater improvements as more information becomes available over time. With additional access to the next clip (\ie $+ 5s$), we have larger improvements of 19\% Avg R@1 and 16\% Cycle@1, with further gains as \methodshort\ is able to find uniqueness over more subsequent clips.

Table \ref{tab:resultstimeloop} shows results on timeloop movies with 2s clips. 
\methodshort\ is able to give a larger number of unique captions when allowed to advance through the story of each repetition. The base model Video-LLaMA  struggles to generate unique captions at longer timescales because it is not conditioned on other clips, and thus has no mechanism to identify uniqueness. Because there is a large amount of information in longer clips, it ends up producing captions based on their most obvious properties, which tend to also be common between clips.
With access to the next 2 clips (\ie $+ 4s$), \methodshort\ improves Avg R@1 by 10\% and Cycle@1 by 19\%, with further gains over more clips.

\vspace{-8pt}
\subsection{Examples}

\begin{figure}[t]
    \centering
    \subfloat[Climbs the stairs\label{fig:egoex1}]{\includegraphics[height=105pt]{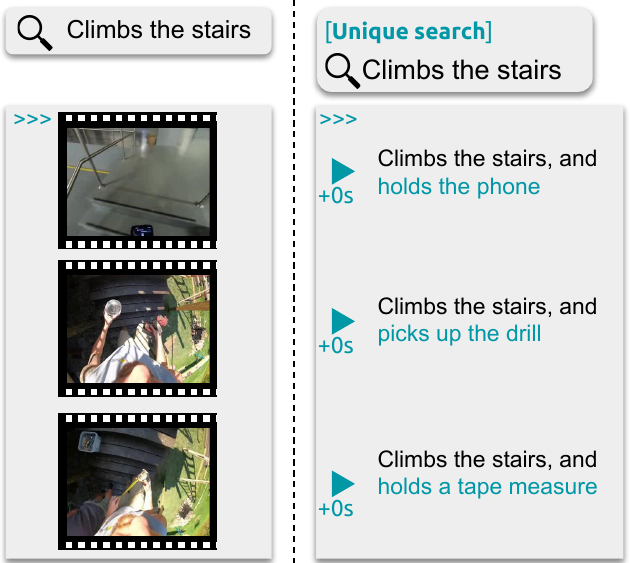}} \hspace{5pt}
    \subfloat[Looks around the shelves.\label{fig:egoex2}]{\includegraphics[height=105pt]{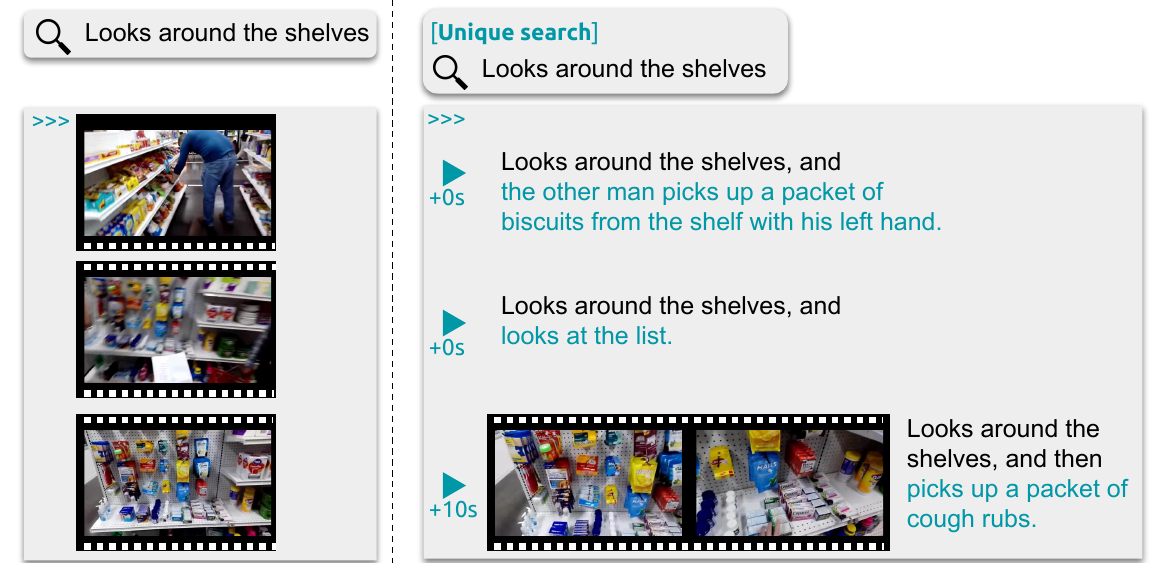}} \\
    \vspace*{-4pt}
    \caption{Qualitative egocentric examples. \methodshort\ is able to caption the set in (a) uniquely. The third clip in (b) advances 10s to generate a unique caption.}
    \vspace*{-6pt}
    \label{fig:egoqual}
\end{figure}

\begin{figure}[t]
    \centering
    \includegraphics[width=\textwidth]{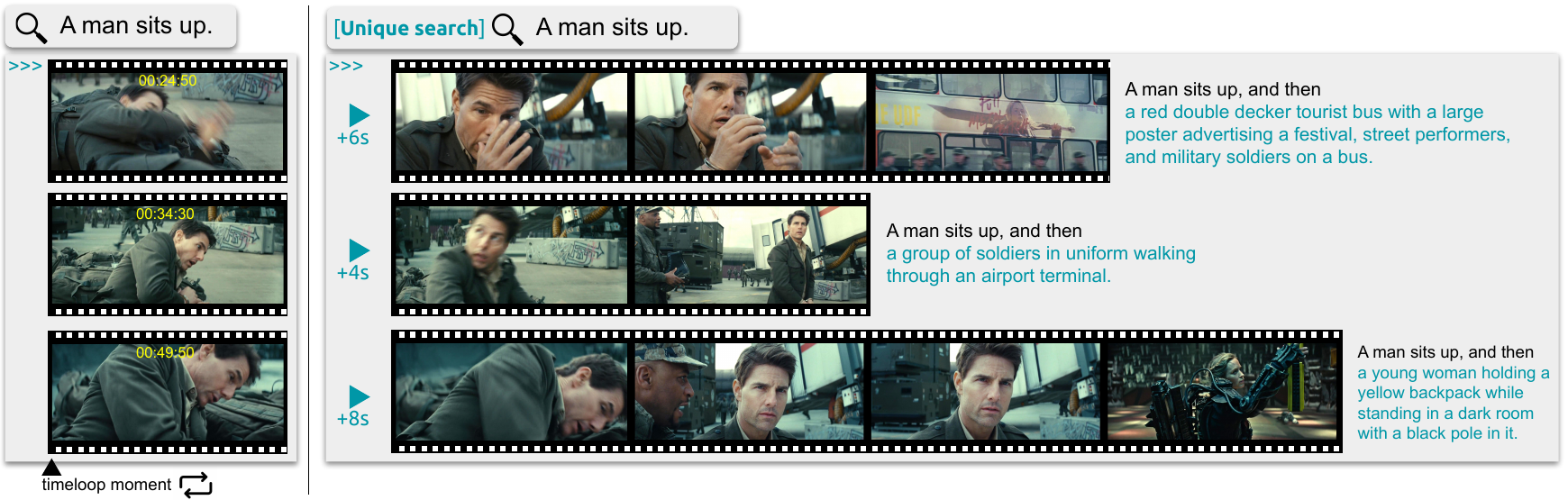}
    \vspace*{-10pt}
    \caption{Unique captioning example on Edge of Tomorrow (2014).}
    \vspace*{-10pt}
    \label{fig:edgeoftomorrow}
\end{figure}

\begin{figure}[t]
    \centering
    \includegraphics[width=\textwidth]{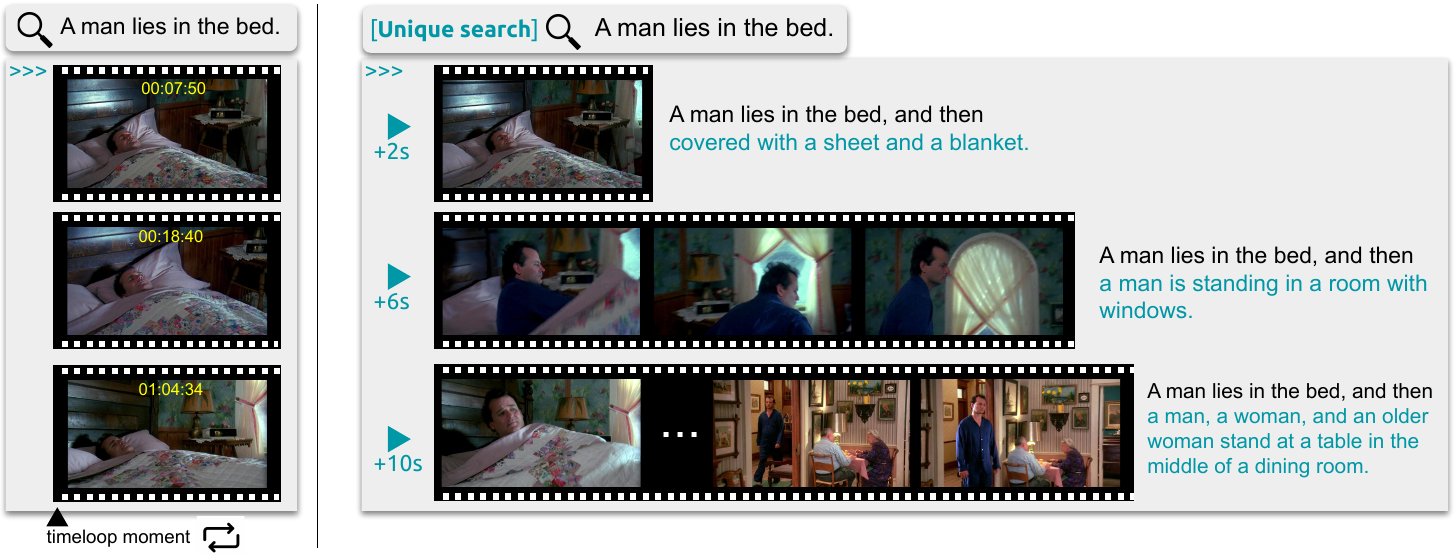}
    \vspace*{-10pt}
    \caption{Unique captioning example on Groundhog Day (1993).}
    \vspace*{-5pt}
    \label{fig:groundhogday}
\end{figure}

Fig. \ref{fig:egoqual} shows qualitative examples on egocentric footage. The unique captions generated by \methodshort\ are displayed to the right of each clip.
\ref{fig:egoex1} shows three clips captioned as ``climbs the stairs''. \methodshort\ is able to predict that all clips can be distinguished by the item the person is holding, and that there is no need to advance through the video.
\ref{fig:egoex2} shows a more challenging example, where the original query is ``looks around the shelves''. \methodshort\ captions the first two clips uniquely. For clip 1, \methodshort\ predicts the \promptname\  ``the other man...'', which conditions the captioner to describe what the other person in the scene is doing. Clip~2 uses the \promptname\  ``looks at...'', indicating the shopping list being read as unique for this clip. 
For clip 3, \methodshort\ cannot find a \promptname\ at +0s or +5s, as captions would also apply to clip 2, and thus would not be unique. At +10s, it predicts uniqueness with the prompt ``picks up''.

Fig. \ref{fig:edgeoftomorrow} shows unique captioning examples on three out of the eight instances of Tom Cruise sitting up in \emph{Edge of Tomorrow}, when a timeloop begins. None of the clips can initially be distinguished. \methodshort\ finds uniqueness using the bus in clip 1 at 6s. In clip 2, the other soldiers in the scene at 4s. In clip 3, at 8s the woman with the backpack is not present in the other clips, resulting in a unique caption. 

Fig. \ref{fig:groundhogday} shows examples on the seminal timeloop movie \emph{Groundhog Day}. We show unique captions generated for three out of the nine loops which start with ``a man wakes up''. At 2s, clip 1 is the only one where the man is still lying down. After 6s, clip 2 is uniquely identified by the objects in the scene (windows). After 10s, \methodshort\ identifies that the other characters and location make it unique.

\begin{figure}[t]
\centering
\begin{minipage}{.75\textwidth}
\centering
\resizebox{8cm}{!}{
\begin{tabular}{@{}lclcrcccl@{}}
\toprule
 & \multicolumn{1}{l}{} &  & \multicolumn{2}{c}{\textbf{Prompts}} & \textbf{V$\rightarrow$T} & \textbf{T$\rightarrow$V} &  &  \\
\textbf{T} & \textbf{\# clips} & \textbf{Method} & \textbf{Max |} & \textbf{Chosen} & \textbf{R@1} & \textbf{R@1} & \textbf{Avg R@1} & \textbf{C@1} \\  \midrule
 &  & {\color[HTML]{999999} LaViLa VCLM} & {\color[HTML]{999999} -} & {\color[HTML]{999999} -} & {\color[HTML]{999999} 45} & {\color[HTML]{999999} 54} & {\color[HTML]{999999} 50} & {\color[HTML]{999999} 34.3} \\
 &  &  & 1 & 1 & 45 & 63 & 54 & 37.2 \\
 &  &  & 2 & 1.4 & 47 & 65 & 56 & 39.6 \\ 
\multirow{-4}{*}{+0s} & \multirow{-4}{*}{1} & \multirow{-3}{*}{LaViLa VCLM + \methodshort} & 3 & 1.6 & \textbf{49} & \textbf{69} & \textbf{59} & \textbf{40.2} \\  \midrule
 &  & {\color[HTML]{999999} LaViLa VCLM} & {\color[HTML]{999999} -} & {\color[HTML]{999999} -} & {\color[HTML]{999999} 51} & {\color[HTML]{999999} 59} & {\color[HTML]{999999} 55} & {\color[HTML]{999999} 39.6} \\
 &  &  & 1 & 1 & 59 & 74 & 67 & 54.0 \\
 &  &  & 2 & 1.5 & \textbf{66} & \textbf{83} & \textbf{75} & \textbf{60.6} \\
\multirow{-4}{*}{+10s} & \multirow{-4}{*}{3} & \multirow{-3}{*}{LaViLa VCLM + \methodshort} & 3 & 1.9 & \textbf{66} & 82 & 74 & 60.4 \\  \midrule
\end{tabular}
}
\vspace{-10pt}
\captionof{table}{Ablation on $\alpha$, the maximum number of prompts. The LaViLa VCLM baseline is shown for comparison.}
\label{tab:maxprompts}
\end{minipage}
\hfill
\begin{minipage}{.19\textwidth}
  \centering
  \includegraphics[scale=0.38, trim= 0 40 0 10]{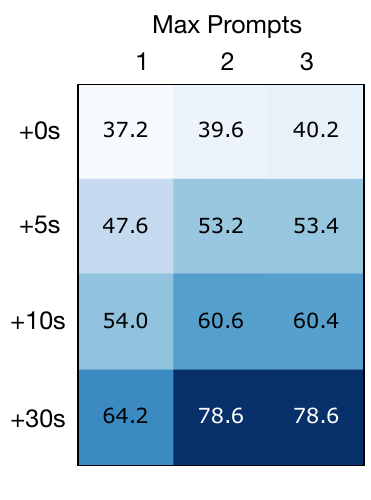}
  \vspace{-10pt}
  \captionof{figure}{Cycle@1 visualisation.}
  \label{fig:maxprompts}
\end{minipage} \hfill
\vspace{-10pt}
\end{figure}

\vspace*{-0pt}
\subsection{Ablations}
\vspace*{-4pt}

Ablations are performed on 50 egocentric sets of 10 clips.

\noindent\textbf{Max allowed prompts $\alpha$.}
We chose the maximum number of prompts per clip as $\alpha=3$ for the main experiments as a reasonable trade-off between caption conciseness and retrieval accuracy. Table \ref{tab:maxprompts} shows results with T = +0s (one clip) and +10s (access to two subsequent clips) as we reduce $\alpha$. We also record the average number of prompts chosen, as not every clip will require as many as $\alpha$ prompts. 
On just the first clip (T = +0s), $\alpha=3$ provides the best results as expected, but with with access to more clips, performance saturates at $\alpha=2$, as visualised in Fig. \ref{fig:maxprompts}. With more time, it is more likely that unique properties will become available. Notably, \methodshort\ with one prompt outperforms the baseline.

\begin{figure}[t]
\centering
\vspace{-6pt}
\subfloat[\% of captions each prompt is chosen for.\label{fig:prompts_used}]{\includegraphics[scale=0.3, trim= -10 30 -10 0]{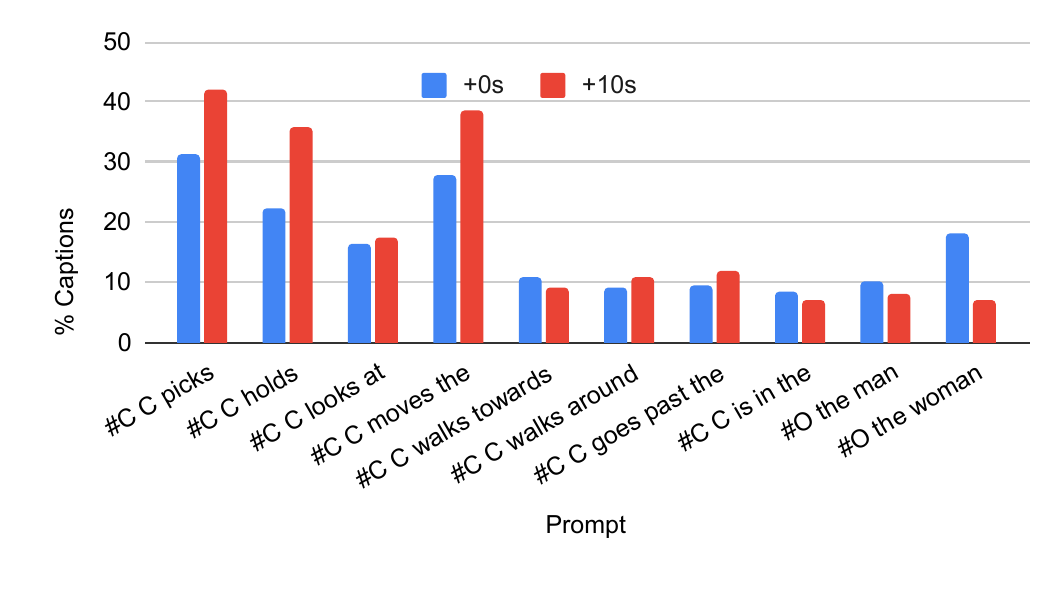}} \hspace{5pt}
\subfloat[Performance of each prompt individually.\label{fig:promptonly}]{\includegraphics[scale=0.3, trim= -10 30 -10 0]{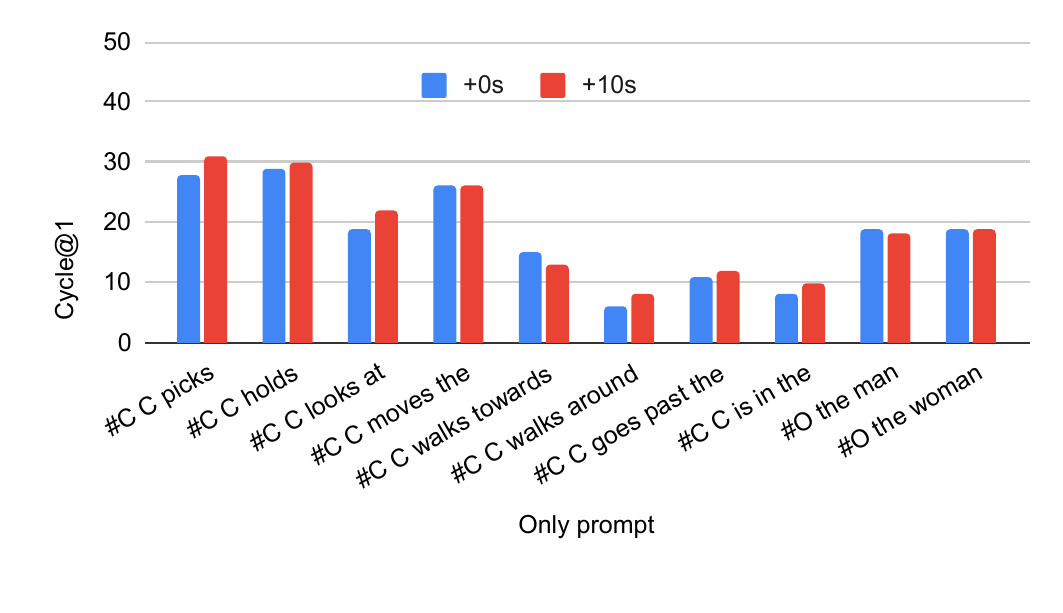}} 
\vspace{-5pt}
\caption{Prompt ablations, given at T= +0s and +10s.}
\vspace{-10pt}
\label{fig:promptindividual}
\end{figure}

\noindent\textbf{Prompt ablation.}\label{sec:ablation_bank}
To investigate the impact of each prompt, \ref{fig:prompts_used} shows when each prompt was chosen (adding up to > 100\% due to combinations being chosen). \ref{fig:promptonly} shows the performance of \methodshort\ with each prompt individually. Prompts relating to active object are chosen the most, and do best individually, as they are frequently the focus of egocentric videos. 

\begin{figure}[t]
\centering
\subfloat[Effect of $\lambda$ on \% clips predicted as unique.\label{fig:lambda_npreds}]{\includegraphics[scale=0.29, trim= -77 0 -30 0]{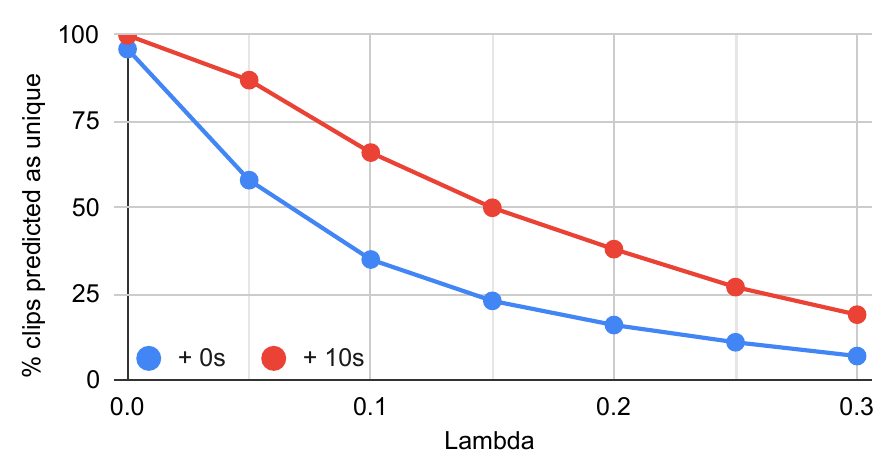}} \hspace{0pt}
\subfloat[Effect of $\lambda$ on Cycle@1 of clips predicted as unique.\label{fig:lambda_c1}]{\includegraphics[scale=0.29, trim= -72 0 -145 0]{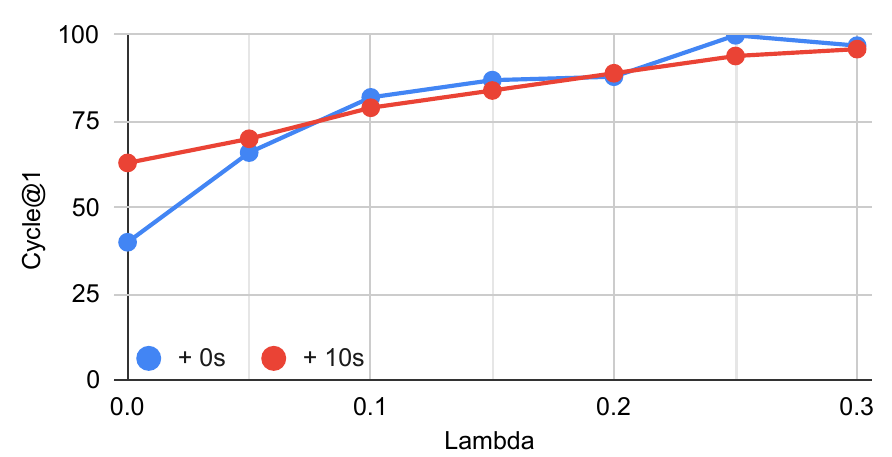}}
\vspace{-8pt}
\caption{Ablation on the margin threshold $\lambda$, on T= +0s and T= +10s. } 
\vspace{-15pt}
\label{fig:lambda}
\end{figure}

\noindent\textbf{Margin threshold $\lambda$.}
in Fig. \ref{fig:lambda_npreds}, we vary $\lambda$, and record the percentage of clips which have a margin $>\lambda$. As expected, a higher $\lambda$ means fewer clips are predicted as unique. Naturally, the number of unique predictions for a fixed $\lambda$ is greater for +10s than +0s, as \methodshort\ has more footage to identify uniqueness.
We measure the Cycle@1 of clips with margin $>\lambda$ in Fig. \ref{fig:lambda_c1}, where higher $\lambda$ gives a higher Cycle@1. 
These results demonstrate that $\lambda$ is a useful parameter to control the decision to advance time, and a proxy for prediction confidence.

\noindent\textbf{Long video case study.}
Instead of our 300K clips, we experiment on one continuous long video, for the task of uniquely captioning every clip in that single video. 
 We demonstrate the effectiveness of the captions produced by 

\begin{wraptable}{r}{5cm}
\centering
\vspace{-15pt}
\resizebox{1.0\linewidth}{!}{
\begin{tabular}{@{}lp{25pt}p{25pt}p{25pt}p{25pt}@{}}
\toprule
\textbf{Method} & \multicolumn{1}{l}{\textbf{R@1}} & \multicolumn{1}{l}{\textbf{R@2}} & \multicolumn{1}{l}{\textbf{R@3}} & \multicolumn{1}{l}{\textbf{R@5}} \\ \midrule
LaViLa VCLM & 12 & 20 & 26 & 33 \\
LaViLa VCLM + \methodshort & \textbf{32} & \textbf{42} & \textbf{48} & \textbf{56} \\ \bottomrule
\end{tabular}
}
\caption{Text $\rightarrow$ Video retrieval on long (40 min) egocentric videos. \label{tab:long_vid}}
\vspace{-20pt}
\end{wraptable}

\noindent\methodshort\ by captioning every 5s clip in 10 long egocentric videos from Ego4D (40 mins and 436 clips on average), and measuring text $\rightarrow$ video recall. 
Tab. \ref{tab:long_vid} shows \methodshort\ delivers significant improvements. Full experimental details are in the supplementary.

\section{Conclusion}

In this paper, we introduced the problem of unique video captioning, to reflect the repetitive nature of daily life, the way repetitions are depicted in film, and the shortcomings of current methods to distinguish between these repetitive events.



We developed a framework, \method\ (\methodshort), based around observing all clips to be captioned. We introduced two benchmarks for unique captioning, based on egocentric footage and the repetitive moments in timeloop movies, and found \methodshort\ provides significant improvements on both. 

There are a number of possible directions for future work. One would be learning prompts vs the fixed set used here. Another would be to explore unique captioning across whole datasets. A third would be to incorporate multiple captioners with different specialisms.

\noindent \textbf{Acknowledgements.} Research is supported by EPSRC Programme Grant Visual AI (EP/T028572/1) and EPSRC UMPIRE (EP/T004991/1). 
This project acknowledges the use of the EPSRC funded Tier 2 facility, JADE-II.

\bibliographystyle{splncs04}
\bibliography{main}


\pagebreak
\appendix

\section{Video examples}

We provide a video on the project's webpage:
\url{http://tobyperrett.github.io/its-just-another-day}, with 6 examples of \method\ (\methodshort) on the timeloop movie and egocentric benchmarks. In each case, we note the matching caption in black and the conditioned caption (by the chosen discriminative prompt(s)) in blue.

\section{Additional models}

In the main paper, we presented results using the SOTA baseline model for each benchmark (egocentric and timeloop), and demonstrated that when incorporating CDP, results improve on both. Here we show Average Recall@1 with other captioners and embedding spaces.

Table \ref{tab:ego_models} shows results on the egocentric benchmark. Note that when evaluating a LaViLa VCLM variant in a LaViLa V/T space, we ensure they are not based on the same model. \ie the default LaViLa V/T space is the Large variant, apart from for the TFS-L LaViLa VCLM, which is evaluated in the Base space. This ensures a model is not evaluated with it's own features for fair comparison. Results in \g{green} indicate those from the main paper.

\begin{table}
\centering
\begin{tabular}{@{}lllll|llll@{}}
\toprule
 & \multicolumn{4}{c}{LaViLa V/T space} & \multicolumn{4}{c}{EgoVLP V/T space} \\ 
Captioner & T=0 & T=5 & T=10 & T=30 & T=0 & T=5 & T=10 & T=30 \\ \midrule
EILEV \cite{yu2023efficient} & 15 & 15 & 15 & 16 & 17 & 17 & 17 & 19 \\
EILEV + CDP & \textbf{17} & \textbf{18} & \textbf{20} & \textbf{26} & \textbf{19} & \textbf{23} & \textbf{26} & \textbf{32} \\ \midrule
TSF-B LaViLa VCLM \cite{zhao2023learning} & 30 & 34 & 34 & 37 & 32 & 34 & 37 & 38 \\
TSF-B LaViLa VCLM + CDP & \textbf{36} & \textbf{48} & \textbf{54} & \textbf{67} & \textbf{37} & \textbf{50} & \textbf{56} & \textbf{67} \\ \midrule
TSF-L LaViLa VCLM \cite{zhao2023learning} & 31 & 36 & 36 & 38 & {\color[HTML]{2a8943} {37}} & {\color[HTML]{2a8943} {38}} & {\color[HTML]{2a8943} {41}} & {\color[HTML]{2a8943} {43}} \\
TSF-L LaViLa VCLM + CDP & \textbf{37} & \textbf{48} & \textbf{54} & \textbf{67} & {\color[HTML]{2a8943} \textbf{{45}}} & {\color[HTML]{2a8943} \textbf{{57}}} & {\color[HTML]{2a8943} \textbf{{65}}} & {\color[HTML]{2a8943} \textbf{{76}}} \\ \bottomrule
\end{tabular}
\caption{Additional models and evaluation spaces on the egocentric benchmark. \label{tab:ego_models}}
\end{table}

Table \ref{tab:timeloop_models} are results on the timeloop movie benchmark. CLIP averages features over all frames. Again, results in \g{green} indicate those from the main paper.

\begin{table}
\centering
\begin{tabular}{@{}lllll|llll@{}}
\toprule
 & \multicolumn{4}{c}{CLIP V/T space} & \multicolumn{4}{c}{InternVideo V/T space} \\ 
Captioner & T=0 & T=2 & T=4 & T=10 & T=0 & T=2 & T=4 & T=10 \\ \midrule
VideoBLIP \cite{Yu_VideoBLIP} & 25 & 24 & 30 & 33 & 33 & 32 & 35 & 36 \\
VideoBLIP + CDP & 25 & 24 & \textbf{35} & \textbf{37} & 33 & \textbf{38} & \textbf{42} & \textbf{50} \\ \midrule
VideoLlama \cite{zhang2023video} & \textbf{31} & 35 & 36 & 33 & {\color[HTML]{2a8943} {35}} & {\color[HTML]{2a8943} {43}} & {\color[HTML]{2a8943} {43}} & {\color[HTML]{2a8943} {38}} \\
VideoLlama + CDP & 28 & \textbf{36} & \textbf{41} & \textbf{42} & {\color[HTML]{2a8943} \textbf{42}} & {\color[HTML]{2a8943} \textbf{{48}}} & {\color[HTML]{2a8943} \textbf{{53}}} & {\color[HTML]{2a8943} \textbf{{63}}} \\ \bottomrule
\end{tabular}
\caption{Additional models and evaluation spaces on the timeloop movie benchmark. \label{tab:timeloop_models}}
\end{table}

CDP delivers larger improvements on better base models. Better base models are more likely to give a correct caption grounded on the visual input when prompted.
This is encouraging, as baseline models will improve over time, and indicates CDP will likely continue to be relevant.

\section{Additional Benchmark Statistics}

\begin{figure}
\centering
\includegraphics[width=0.8\textwidth]{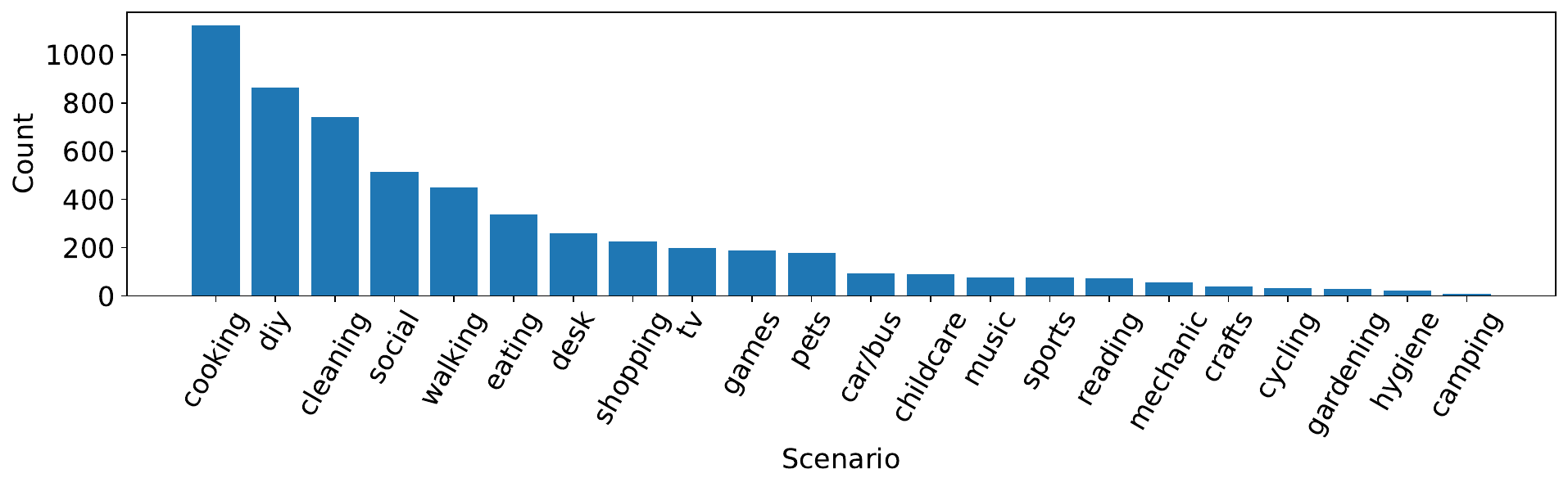}
\caption{Scenarios in the egocentric benchmark.}
\label{fig:scenario_counts}
\end{figure}

\begin{figure}
\centering
\includegraphics[width=0.6\textwidth]{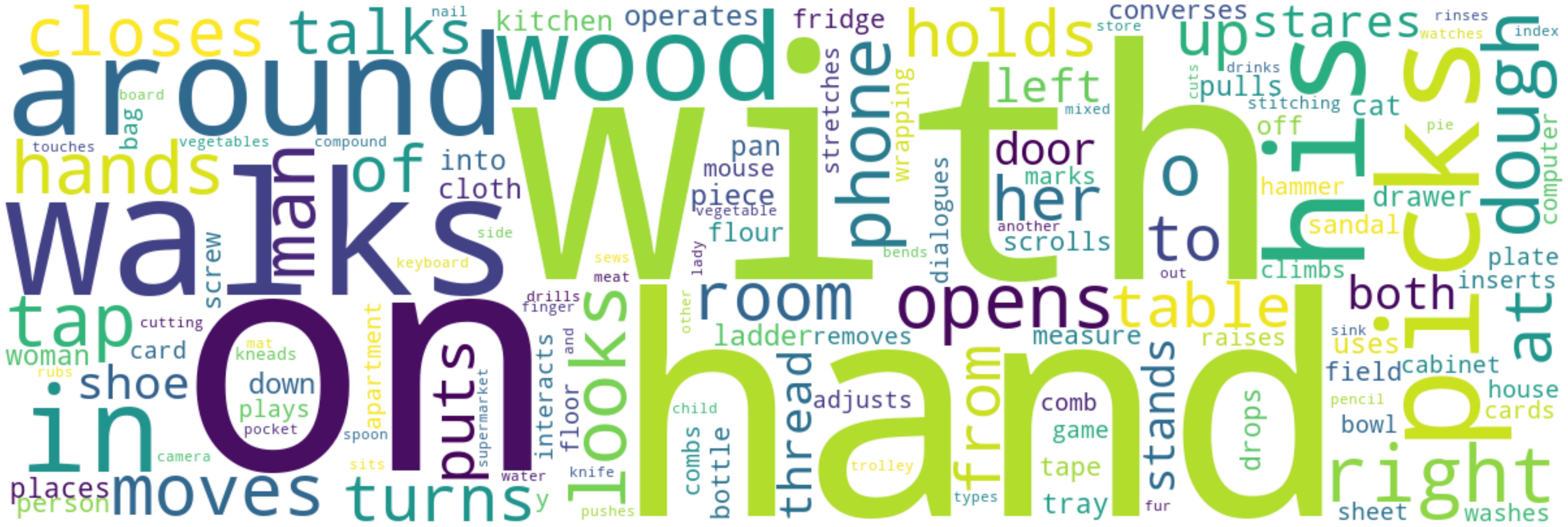}
\caption{Wordle of narrations used to create the egocentric benchmark.}
\label{fig:wordle}
\end{figure}

Fig. \ref{fig:scenario_counts} explores the distribution of scenarios which appear in the egocentric benchmark. These roughly match the scenarios present in Ego4D. We also show a wordle of the narrations used to generate the benchmark in Fig. \ref{fig:wordle}. Interestingly, movement features a lot (\eg ``walks'', ``around''), and it is often difficult for models to distinguish between different parts of an environment.

\section{Case Study: Long Egocentric Text-to-Video Retrieval}

In the paper, we evaluated unique captioning on sets of 10 identical narrated clips drawn from Ego4D, and showed that 
\methodshort\ is able to significantly improve the retrieval performance of the LaViLa VCLM on this task. The final ablation is expanded here, where we give an example retrieval use case on long egocentric videos. 

\begin{figure}[t!]
\centering
\includegraphics[width=\textwidth]{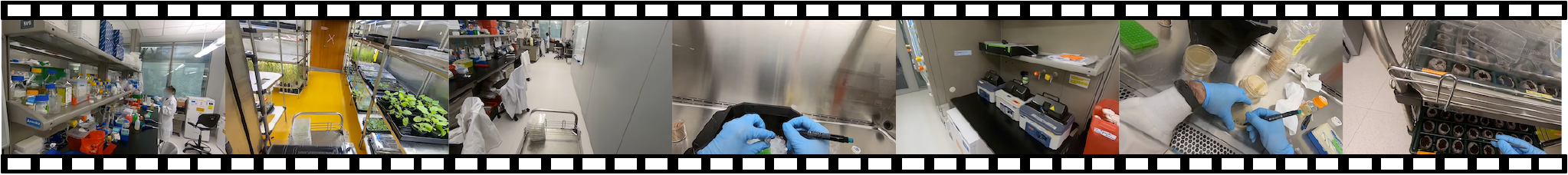}
\caption{Example of a long 36 minute egocentric video in a lab.}
\label{fig:egolab}
\end{figure}

We select 10 long videos from the Ego4D NLQ test set from different scenarios (lab work, cooking, sports, construction \etc). An example is shown in Figure \ref{fig:egolab}. We break each video into consecutive 5s clips (\ie clips are 0-5s, 5-10s, 10-15s...). The videos have an average length of 40.3 minutes, containing 483 clips each on average.

When attempting to caption, some clips will be similar producing identical captions. Temporally consecutive clips are especially challenging. We demonstrate how \methodshort\ can be used to improve retrieval with better captions, resulting in more effective text-to-video retrieval. 

\subsection{Experiment}
We assess unique caption quality on the long video with Text$\rightarrow$Video retrieval.
We perform Text$\rightarrow$Video retrieval in the joint video/text embedding space, where a text embedding is used as a query, and the result is the video with the closest embedding. 

For each clip, we generate its caption using either (i) LaViLa VCLM alone, or (ii) LaViLa VCLM with \methodshort. These captions are the text queries. We then attempt to retrieve each video clip by its generated caption in the shared video/text space, and measure Text$\rightarrow$Video R@1, R@2, R@3 and R@5 retrieval. This is a good test of unique captioning, as better captions will obtain higher retrieval scores due to less confusion with clips they are not generated from. If a clip is not uniquely captioned, then multiple captions could refer to a single clip, giving lower retrieval scores.

Both methods have access to T = +5s (\ie the clip plus one subsequent clip). Note that we allow LaViLa VCLM to view both clips at once, as in the main experiments (as this performs better than just one clip).

\begin{table}[t]
\centering
\begin{tabular}{@{}lp{25pt}p{25pt}p{25pt}p{25pt}@{}}
\toprule
\textbf{Method} & \multicolumn{1}{l}{\textbf{R@1}} & \multicolumn{1}{l}{\textbf{R@2}} & \multicolumn{1}{l}{\textbf{R@3}} & \multicolumn{1}{l}{\textbf{R@5}} \\ \midrule
LaViLa VCLM & 12 & 20 & 26 & 33 \\
LaViLa VCLM + \methodshort & \textbf{32} & \textbf{42} & \textbf{48} & \textbf{56} \\ \bottomrule
\end{tabular}
\caption{Text $\rightarrow$ Video retrieval on long egocentric videos (average 40 minutes). \label{tab:long_tvr}}
\end{table}

\subsection{Results}
Table \ref{tab:long_tvr} shows Text $\rightarrow$ Video R@1, R@2 and R@3.
\methodshort\ obtains an R@1 improvement of 21.5\% compared to the LaViLa VCLM (36.0\% compared to 14.5\%), with larger gains for R@2 (+28.1\%) and R@3 (+28.5\%). Interestingly, \methodshort\ R@1 is higher than LaViLa R@3.

\subsection{Complexity}
In Section 3.2 of the main paper, we discussed the complexity of the search. 
For a 40 minute video, using $\alpha=3$ and 5s clips, the exact combinatorial search requires $<1s$ on one CPU core.
Even for a video 10x this length (6 hours), the search would take $<30s$ on one CPU core, and is embarrassingly parallel.
Our code is publicly available from the project's webpage.

\section{Accuracy of \methodnet}

\begin{figure}
    \centering
    \includegraphics[width=0.7\textwidth]{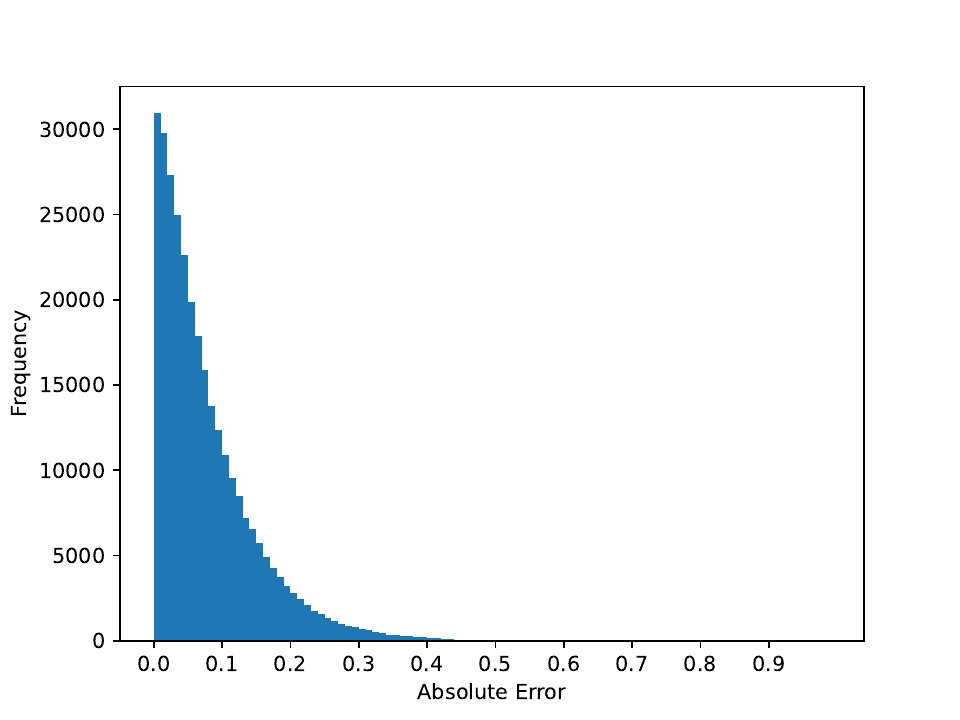}
    \caption{Error of \methodnet\ on a held out validation set.}
    \label{fig:error}
\end{figure}
Figure \ref{fig:error} shows a histogram of the absolute errors of \methodnet, on a held out validation set of egocentric footage, containing 30,000 clip/caption pairs. The absolute error is the mean of the absolute difference between the ground-truth video/caption cosine similarity (Eq. 2 in main paper), and the predicted similarity by \methodnet\ (Eq. 6 in main paper):
\begin{equation}
    \text{absolute error} = | \hat s - s |
\end{equation}
The figure shows most errors to be small, and the error has mean = 0 and standard deviation = 0.11.

\section{Prompts}
The 10 prompts used for the egocentric benchmark (ablated in Section 5.5) are:
\begin{itemize}
\item \#C C picks
\item \#C C holds
\item \#C C looks at
\item \#C C moves the
\item \#C C walks towards the
\item \#C C walks around the
\item \#C C goes past the
\item \#C C is in the
\item \#O the man
\item \#O the woman
\end{itemize}

\noindent The 10 prompts used for timeloop movies are:
\begin{itemize}
\item Who is in the scene in this video?
\item What is the man doing in this video?
\item What is the woman doing in this video?
\item Where are they in this video?
\item What are they picking in this video?
\item Who are they talking with in this video?
\item What are they holding in this video?
\item What are they looking at in this video?
\item What are they moving in this video?
\item Where are they going in this video?
\end{itemize}

\end{document}